\documentclass[final]{cvpr}
\usepackage{times}
\usepackage{epsfig}
\usepackage{graphicx}
\usepackage{amsmath}
\usepackage{amssymb}
\usepackage{booktabs} 
\usepackage{algorithm}
\usepackage{algorithmic}
\usepackage{amsthm}
\usepackage{diagbox}
\usepackage{subfigure}
\usepackage{caption}
\usepackage{float}
\usepackage{epstopdf}
\usepackage{enumitem}
\usepackage{multirow}
\usepackage{bbm}

\usepackage[pagebackref=true,breaklinks=true,colorlinks,bookmarks=false]{hyperref}

\def\cvprPaperID{2548} 
\def\confYear{CVPR 2021}
\begin{document}

\title{I$^3$Net: Implicit Instance-Invariant Network for Adapting \\ One-Stage Object Detectors}

\author{Chaoqi Chen$^1$, Zebiao Zheng$^2$, Yue Huang$^{2*}$, Xinghao Ding$^2$, Yizhou Yu$^{1,3}\thanks{Corresponding authors}$\\
    {$^1$~The University of Hong Kong}\\
	{$^2$~School of Informatics, Xiamen University, China}\\
    {$^3$~Deepwise AI Lab}\\
     
	{\tt\small cqchen1994@gmail.com, zbzheng@stu.xmu.edu.cn}\\
	{\tt\small huangyue05@gmail.com, dxh@xmu.edu.cn, yizhouy@acm.org}
}

\maketitle
\pagestyle{empty}
\thispagestyle{empty}

\begin{abstract} 
Recent works on two-stage cross-domain detection have widely explored the local feature patterns to achieve more accurate adaptation results. These methods heavily rely on the region proposal mechanisms and ROI-based instance-level features to design fine-grained feature alignment modules with respect to the foreground objects. However, for one-stage detectors, it is hard or even impossible to obtain explicit instance-level features in the detection pipelines. Motivated by this, we propose an Implicit Instance-Invariant Network (I$^3$Net), which is tailored for adapting one-stage detectors and implicitly learns instance-invariant features via exploiting the natural characteristics of deep features in different layers. Specifically, we facilitate the adaptation from three aspects: (1) Dynamic and Class-Balanced Reweighting (DCBR) strategy, which considers the coexistence of intra-domain and intra-class variations to assign larger weights to those sample-scarce categories and easy-to-adapt samples; (2) Category-aware Object Pattern Matching (COPM) module, which boosts the cross-domain foreground objects matching guided by the categorical information and suppresses the uninformative background features; (3) Regularized Joint Category Alignment (RJCA) module, which jointly enforces the category alignment at different domain-specific layers with a consistency regularization. 
Experiments reveal that I$^3$Net exceeds the state-of-the-art performance on benchmark datasets.
\end{abstract}
\begin{figure}[!t]
\centering
\setlength{\belowcaptionskip}{-0.5cm}
\includegraphics[width=0.44\textwidth]{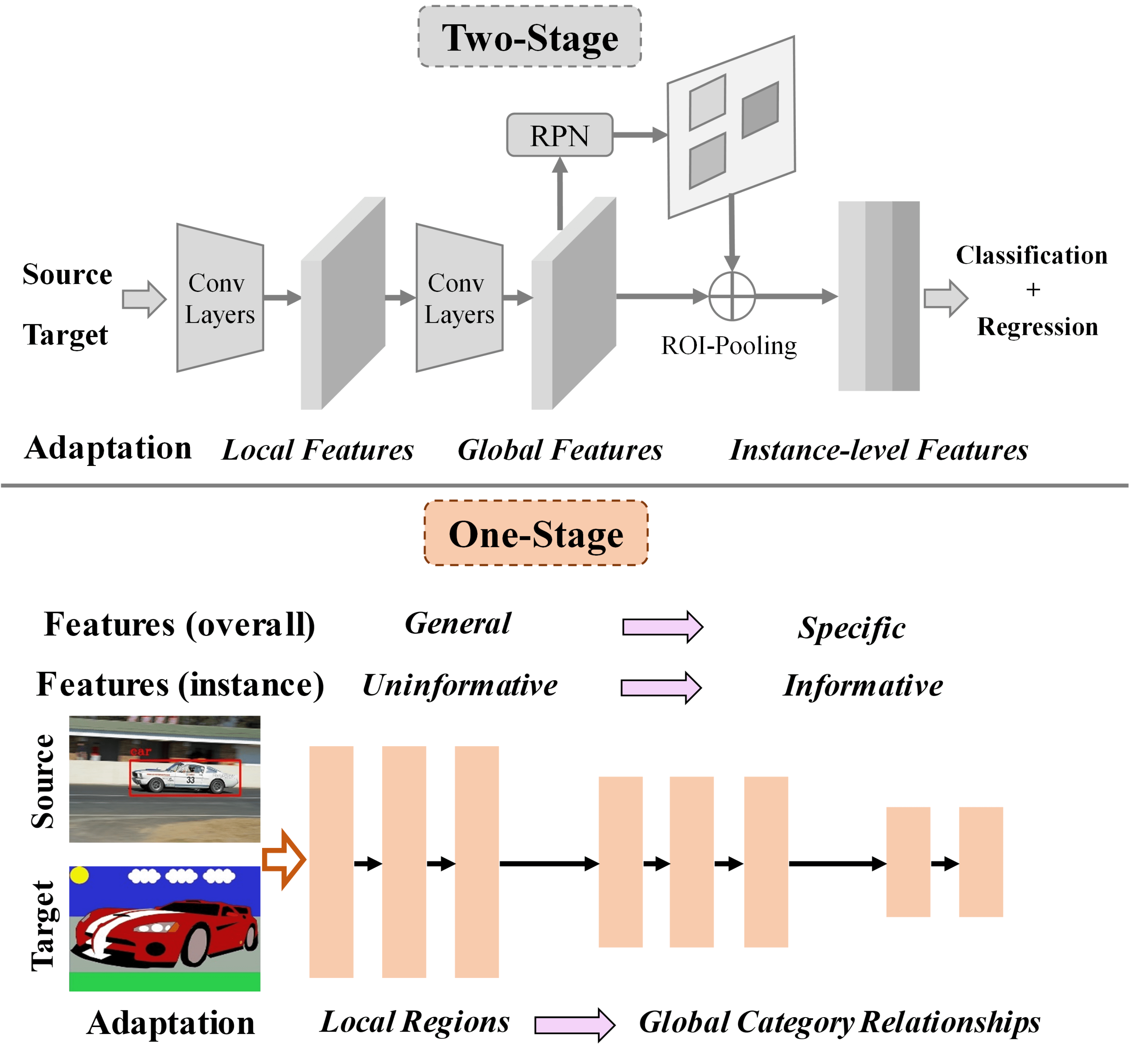}
\vspace{-0.2cm}
\caption{\textbf{Upper:} Illustration of previous two-stage cross-domain detection methods. \textbf{Lower:} Motivation of the proposed method based on the observation with respect to the characteristics of deep features in different layers.}\label{fig1}
\end{figure}
\section{Introduction}
\label{sec:intro}
Object detection has achieved remarkable progress due to the unprecedented development of deep convolutional networks (CNNs) and the existence of large-scale annotated datasets. However, collecting large amounts of instance-level annotated data in various domains for object detection is prohibitively costly. An alternative would be applying the off-the-shelf detection model trained on the source domain to a new target domain. However, deep object detectors suffer from performance degradation when applied to a new domain under the presence of domain shift~\cite{torralba2011unbiased}.
This problem has inspired the research on Unsupervised Domain Adaptation (UDA)~\cite{pan2010survey}, which aims to bridge the distribution discrepancy between source and target domains via knowledge transfer. 
Numerous approaches, such as moment matching~\cite{gong2012geodesic,fernando2013unsupervised,long2015learning,long2017deep,zellinger2017central} and adversarial learning~\cite{ganin2015unsupervised,tzeng2017adversarial,shu2018dirt,long2018conditional,xie2018learning}, have been proposed for cross-domain image classification and semantic segmentation.

Compared to the conventional UDA problems, cross-domain object detection is a more sophisticated and challenging problem since the adaptation of classification and regression should be simultaneously considered. Current methods~\cite{chen2018domain,zhu2019adapting,saito2019strong,cai2019exploring,He_2019_ICCV,chen2020harmonizing,xu2020cross,zheng2020cross,xu2020exploring} mostly resort to the adversarial feature adaptation to explore discriminative feature patterns at local-level, global-level, and instance-level for adapting two-stage detectors (see top of Fig.~\ref{fig1}), Faster R-CNN~\cite{ren2015faster}. 
However, they heavily rely on the region proposal mechanisms and ROI-based instance-level features to design fine-grained feature alignment modules with respect to the foreground objects. 
For example, Zhu~\emph{et al.}~\cite{zhu2019adapting} mine the target discriminative regions based on the region proposals derived from the RPN. Cai~\emph{et al.}~\cite{cai2019exploring} regularize the relational graphs by using the ROI-based features. Chen~\emph{et al.}~\cite{chen2020harmonizing} and Xu~\emph{et al.}~\cite{xu2020exploring} assist the instance-level feature alignment by the contextual or categorical regularization.

One-stage object detectors, such as SSD~\cite{liu2016ssd} and RetinaNet~\cite{lin2017focal}, have the merits of being faster and simpler in real-world applications. Unfortunately, it is unrealistic to obtain \emph{explicit} instance-level features in the one-stage detectors due to the lack of region proposal step. Hence, how to adapt one-stage detectors is vital for practical scenarios but yet to be thoroughly studied. 
The motivation of this paper is shown in the bottom of Fig.~\ref{fig1}. Deep features in the standard CNNs must eventually transition from general to specific along the network~\cite{yosinski2014transferable}. Inspired by this, in one-stage detectors, 
we can reasonably envision that the features at lower layers (\emph{e.g.,} color, corner, edge, and illumination) are expected to be 
mostly \emph{instance-uninformative}, while the features at higher layers (\emph{e.g.,} object categories) are \emph{instance-informative}. 
Therefore, we need to alleviate the negative influence of uninformative features and promote the alignment of informative features, \emph{i.e.,} suppress redundant~(such as background) information from the lower layers and enhance the cross-domain semantic correlation of foreground objects at the higher layers.

In this paper, we propose an Implicit Instance-Invariant Network (I$^3$Net) that removes the need for requiring explicit instance-level features. 
Instead, we implicitly learn instance-invariant features via the alignment of transferable regions and images while preserving the inter-domain class relationships. 
To be specific, we facilitate the adaptation of one-stage detectors from three aspects. 
Firstly, upon observing that there exist two conceptually orthogonal distribution variations hidden in the target data, \emph{i.e.,} intra-domain and intra-class variations, 
we propose a Dynamic and Class-Balanced Reweighting (DCBR) strategy to dynamically reweight each target sample based on its \emph{adaptation difficulty}, which is measured by the degree of class imbalance and the prediction uncertainty of a multi-label classifier. 
Secondly, considering that object with the same category label but from different domains will share similar object patterns, we design a Category-aware Object Pattern Matching (COPM) module to boost cross-domain foreground objects matching guided by the categorical information and suppress the uninformative background features at lower layers. 
Finally, we develop a Regularized Joint Category Alignment (RJCA) module to enable category alignment by considering complementary effect of different domain-specific layers and further incorporate a consistency regularization term with respect to the average prediction of different detection heads. 
Experimental results show that the proposed I$^3$Net significantly improves the state-of-the-art performance of one-stage cross-domain object detection on three benchmarks. 

\begin{figure*}[!t]
\centering
\setlength{\belowcaptionskip}{-0.2cm}
\includegraphics[width=0.92\textwidth]{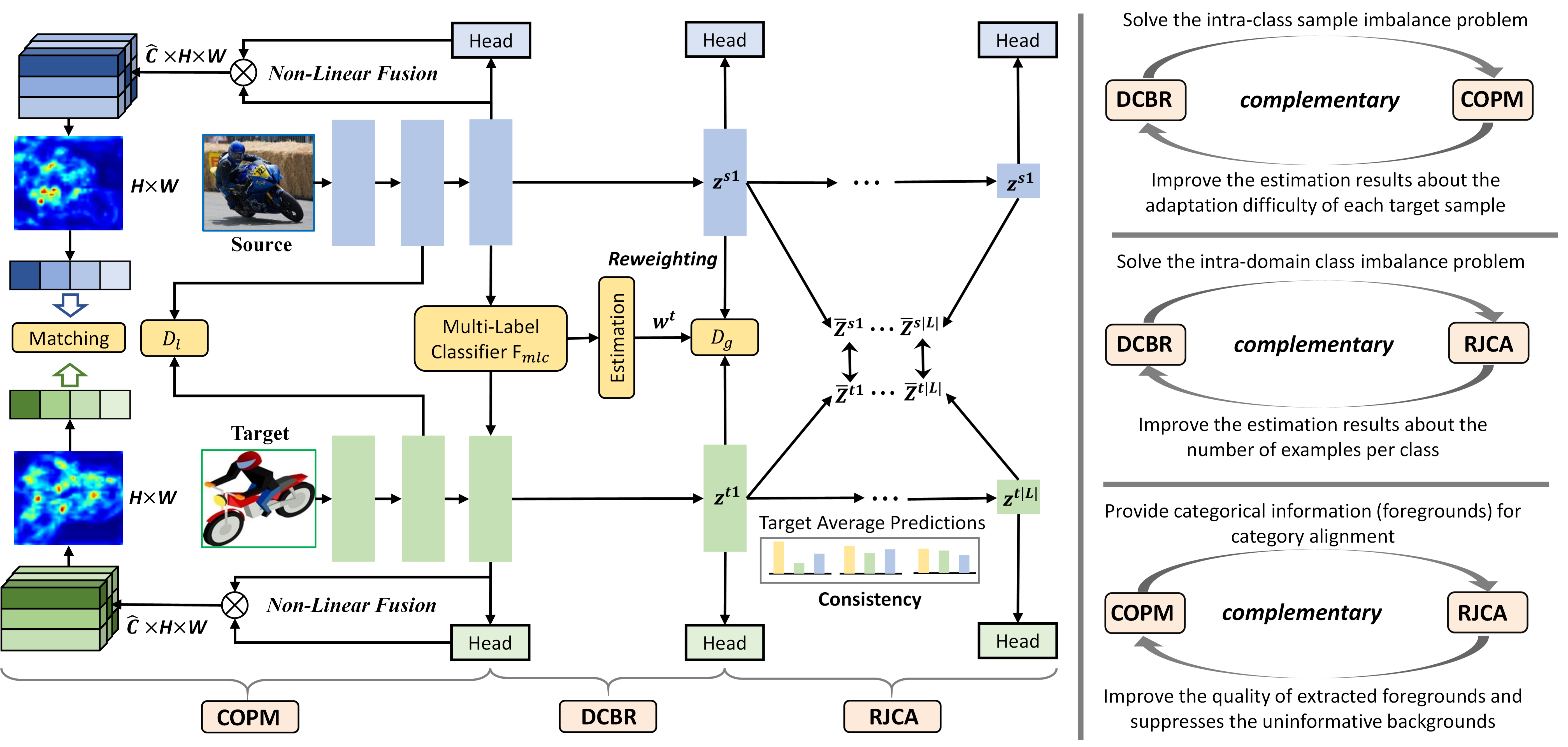}
\vspace{-0.2cm}
\caption{\textbf{Left:} The overall structure of the proposed I$^3$Net, where $F_{mlc}$ is an image-level multi-label classifier, $D_l$ and $D_g$ are pixel-level and image-level domain discriminators respectively. Non-linear fusion stands for the tensor product operation. We adopt SSD as the base detection network. \textbf{Right:} DCBR, COPM, and RJCA are complementary to each other. 
}\label{fig2}
\end{figure*}
\section{Related Work}
\paragraph{Unsupervised Domain Adaptation~(UDA)}
UDA methods have attracted much attention for alleviating the distributional variations between two distinct domains in image classification, semantic segmentation, and object detection.
For UDA, a typical solution is to match the source and target feature distributions in the common space by embedding disparity measures into deep architectures, such as Maximum Mean Discrepancy (MMD)~\cite{tzeng2014deep,long2015learning}, Correlation Alignment (CORAL)~\cite{sun2016deep}, Central Moment Discrepancy (CMD)~\cite{zellinger2017central}, and transport distance~\cite{li2020enhanced,xu2020reliable}. Inspired by the success of Generative Adversarial Nets (GAN)~\cite{goodfellow2014generative}, a large amount works~\cite{ganin2016domain,tzeng2017adversarial,saito2018maximum,pei2018multi,xie2018learning,Chen_2019_CVPR,zhang2019domain,jiang2020implicit} have been done by adversarially learning domain-invariant representations with extra categorical regularization. 
\vspace{-0.35cm}
\paragraph{Object Detection}
Object Detection is one of the most fundamental computer vision problems in the past few decades~\cite{zou2019object}. Our work focuses on how to adapt object detectors, so we only review several representative two-stage and one-stage detectors. The series of region-based convolutional networks~(\emph{i.e.,}~R-CNN~\cite{girshick2014rich}, Fast R-CNN~\cite{girshick2015fast}, and Faster R-CNN~\cite{ren2015faster}) have achieved compelling results in terms of detection accuracy. They count on the region proposal mechanisms to classify region of interest (ROI) independently~\cite{girshick2014rich}, or share the convolution features with ROI pooling layer~\cite{girshick2015fast}, or produce the region proposals based on a Region Proposal Network (RPN)~\cite{ren2015faster}. On the other hand, one-stage detectors, such as SSD~\cite{liu2016ssd}, YOLO~\cite{redmon2016you,redmon2017yolo9000}, and RetinaNet~\cite{lin2017focal} have shown a clear superiority on the inference speed by directly carrying out the category confidence prediction and the bounding box regression. 
\vspace{-0.35cm}
\paragraph{UDA for Object Detection}
Domain Adaptive Faster R-CNN~\cite{chen2018domain} is a pioneering two-stage cross-domain detection method that reduces the distributional shift by adversarially learning domain-invariant features on both image-level and instance-level. Considered the local nature of object detection task, most recent efforts~\cite{zhu2019adapting,saito2019strong,cai2019exploring,He_2019_ICCV,chen2020harmonizing,xu2020cross,zheng2020cross,xu2020exploring,hsu2020every,su2020adapting,zhao2020collaborative} are devoted to capture the local feature patterns and explicitly align them at multiple levels. For instance, 
Chen~\emph{et al.}~\cite{chen2020harmonizing} propose to hierarchically calibrate the transferability of different level features (\emph{i.e.,} local-region, image, and instance) to improve the discriminability of detectors; Xu~\emph{et al.}~\cite{xu2020cross} and Zheng~\emph{et al.}~\cite{zheng2020cross} draw motivation from the cross-domain prototype alignment~\cite{xie2018learning,Chen_2019_CVPR,pan2019transferrable} to align the foreground objects with the same category between domains. However, these methods can not be simply extended to the one-stage detectors since they highly rely on the region proposals and pooled instance-level features. The study on adapting one-stage object detectors is very limited. A pioneering attempt~\cite{kim2019self} present a weak self-training strategy by simultaneously reducing the false positives and false negatives during the hard negative mining. However, self-training-based method may be vulnerable to the error accumulation problem, especially on the sophisticated cross-domain detection scenarios. Moreover, how to learn instance-invariant feature representations without the guidance of region proposal mechanisms, which is crucial for adapting one-stage detectors, still remains unclear. 
\section{Methodology}
In the task of cross-domain object detection, we are given a source domain $\mathcal{D}_s=\{({x_i^s},{y_i^s},{b_i^s})\}_{i=1}^{N_s}$ (${y_i^s}\in{\mathcal{R}^{k\times{1}}}$, $b_i^s\in\mathcal{R}^{k\times{4}}$) of $N_s$ labeled samples, and a target domain $\mathcal{D}_t \! = \! \{{x_j^t}\}_{j=1}^{N_t}$ of $N_t$ unlabeled samples. $\mathcal{D}_s$ and $\mathcal{D}_t$ are drawn from different data distributions, but share an identical group of classes ($K$ classes in all). The objective of this paper is to transfer knowledge from $\mathcal{D}_s$ to $\mathcal{D}_t$ and achieve good detection results in $\mathcal{D}_t$. 
\vspace{-0.3cm}
\paragraph{Framework Overview.}
To this end, we propose an Implicit Instance-Invariant Network (I$^3$Net), which is comprised of three components, namely, Dynamic and Class-Balanced Reweighting (DCBR), Category-aware Object Pattern Matching (COPM), and Regularized Joint Category Alignment (RJCA). The overview of I$^3$Net is demonstrated in Figure~\ref{fig2}. The basic idea is to utilize the inherent characteristics of representations at different layers of the detector to compensate for the lack of explicit instance-level features. 
DCBR reweights target samples based on the adaptation difficulty with respect to the intra-domain and intra-class variations, COPM captures the foreground object patterns and suppresses redundant background information, and RJCA promotes the cross-domain category alignment in different domain-specific layers (connected with detection heads) with a consistency regularization. Following the prior work on adapting one-stage detectors~\cite{kim2019self}, our I$^3$Net is based on the SSD~\cite{liu2016ssd} framework. 

\subsection{Dynamic and Class-Balanced Reweighting}
\label{sec:DCBR}
To date cross-domain detection methods~\cite{chen2018domain,saito2019strong,xu2020cross,zheng2020cross} mainly focus on the feature-level adaptation and treat all the target samples equally, while they neglect the distributional characteristics of the target data, which are crucial for the adaptation process. 
By contrast, the proposed DCBR strategy explicitly explores the intra-domain and intra-class variations within the unlabeled target domain to assign larger weights to those sample-scarce categories and easy-to-adapt samples. We analysis these two variations in the following. 

\textbf{Intra-Domain Variations.} The class-imbalance problem~\cite{oksuz2020imbalance}, which refers to the inequality among the number of examples belonging to different classes, commonly exists in the object detection. Prior efforts, such as Focal Loss~\cite{lin2017focal} and hard example mining~\cite{liu2016ssd,shrivastava2016training}, are devoted to tackle the \emph{foreground-background class imbalance}, which is irrelevant to the number of examples per class in a single domain. In cross-domain detection, we argue that the \emph{foreground-foreground class imbalance}, which is dataset-relevant and may be different between domains, 
is prone to deteriorate the adaptation performance since the adaptation of each class will be affected by the number of examples per class in both domains, \emph{i.e.,} the adaptation difficulty of different categories may be distinct. 

\textbf{Intra-Class Variations.} Owing to the difference of background, object co-occurrence, and scene layouts across domains, 
excessively align the source and target features in the full dataset will result in negative transfer, \emph{i.e.,} some target samples may be less transferable or even non-transferable.
However, most leading cross-domain detection methods treat the target domain as a whole without considering the structures of intra-class data distributions. Motivated by this, we assume that the adaptation difficulty of different samples within the same class may be distinct. An intuitive solution is to utilize re-weighting techniques.
However, this solution suffers a critical limitation in the context of cross-domain object detection. In contrast to the classification problem, where a single image usually contains only one semantic category, there exist multiple instances within the same image in the detection problem. Thus, how to measure the adaptation difficulty of an input target sample in cross-domain detection remains unclear.

Based on the above discussions, we formally provide the details of the proposed DCBR. The DCBR consists of two steps: 
(i) \emph{estimate the adaptation difficulty of each individual target sample and each target class}; 
(ii) \emph{reweight target samples based on the estimation results}. 
Technically, the adaptation difficulty of an target example $x^t$ (for ease of denotation, we omit the subscript of $x_i^s$ and $x_j^t$ when they apply) is measured by using an image-level multi-label classifier (\emph{i.e.,} $F_{mlc}$ in Fig.~\ref{fig2}). 
We first train $F_{mlc}$ based on the labeled source samples for initialization. The multi-label classification loss on the source domain is formulated as:
\begin{equation}\label{eq:mlc}
\mathcal{L}_{\text{mlc}}=\sum_{k=1}^{K}y_k^s\cdot\log(\hat{y}_k^s)+(1-y_k^s)\cdot\log(1-\hat{y}_k^s)
\end{equation}
where $y_k^s$ is the $k^{\text{th}}$ ($k=\{1,2,...,K\}$) element of $y^s$ and $\hat{y}_k^s=F_{mlc}(G_1(x^s))^k$ ($G_1$ is a feature extractor connected to $F_{mlc}$). $y_k^s=1$ means that there exists at least one object of class $k$ in $x^s$; otherwise, $y_k^s=0$ indicates that $x^s$ does not contain the object of class $k$. For each target sample $x^t$, we denote the prediction of its multi-label classification as $\hat{y}_k^t=F_{mlc}(G_1(x^t))^k$. Then, we define the weight function of a target sample $x^t$ \emph{w.r.t.} the intra-class variation by using its multi-label classification output:
\begin{equation}\label{eq:w1}
w_1^t=\frac{1}{K'}\sum_{k=1}^{K}\mathbbm{1}(\hat{y}_k^t(x^t)>\tau)\cdot\hat{y}_k^t(x^t)+1
\end{equation}
where $K'=\sum_{k=1}^{K}\mathbbm{1}(\hat{y}_k^t(x^t)>\tau)$, and $\tau$ is a threshold. $\mathbbm{1}(a)$ is an indicator function which is 1 if $a$ is true and 0 otherwise. By doing so, target samples with higher classification confidence scores will be assigned larger weights since they are more similar with source domain. Note that the value of $w_1^t$ increases continuously since the source and target distributions are getting closer as training proceeds. 

To estimate the number of examples per class in $\mathcal{D}_t$, we resort to the classification output for roughly dividing $\mathcal{D}_t$ into $K$ classes. $x^t$ is added into the target domain of the class $\mathcal{D}_t^{k'}$ if $k' = \arg \max\limits_{k}\; \hat{y}_t^k(x^t)$. Then, the unlabel target samples $\mathcal{D}_t$ are split into $K$ classes, \emph{i.e.,} $\mathcal{D}_t=\{\mathcal{D}_t^k\}_{k=1}^K$. To this end, we are able to assign larger weights to those sample-scarce categories. The weight function of $x^t$ \emph{w.r.t.} the intra-domain variation is formulated as,
\begin{equation}\label{eq:w2}
w_2^t=e^{(1-{N_t^k}/{N_t})}
\end{equation}
where $N_t^k$ denotes the number of samples in class $k$.

Based on Eq.~\eqref{eq:w1} and Eq.~\eqref{eq:w2}, the overall weight function of a target sample $x_j^t$ is formulated as follows,
\begin{equation}\label{eq:w}
w^t=\theta{w_1^t}+(1-\theta)w_2^t
\end{equation}
where $\theta$ is a hyper-parameter to balance $w_1^t$ and $w_2^t$. After adding the weights to all target samples, the adversarial loss of image-wise domain discriminator $D_{g}$ can be written as: 
\begin{equation}\label{eq:ga}
\footnotesize
\begin{split}
\mathcal{L}_{\text{dcbr}}=&-\frac{1}{N_s}\sum\limits_{i=1}^{N_s}\log(D_g(G_2(x_i^s))) \\
&-\frac{1}{N_s}\sum\limits_{j=1}^{N_t}w_j^t\cdot\log(1-D_g(G_2(x_j^t)))
\end{split}
\end{equation}
where $G_2$ is a feature extractor that is connected to $D_g$. 

\subsection{Category-Aware Object Pattern Matching}
As we discussed in Section~\ref{sec:intro}, the feature representations at lower layers contain various redundant information (\emph{e.g.} background) and should not be fully aligned. Previous works~\cite{chen2018domain,saito2019strong}, which strictly matching the low-level features, may result in inferior performance especially on the one-stage detection. During the exploration, we observe that objects with the same category label but from different domains will own similar object patterns. Object pattern, which refers to the discriminative features of foreground objects, can provide rich semantic information \emph{w.r.t.} the objects, such as object category, shape, size, \emph{etc}. Driven by this finding, we propose a Category-aware Object Pattern Matching (COPM) module to boost cross-domain foreground objects matching guided by the categorical information and suppress the uninformative background features. 

Suppose that we have a CNN layer (\emph{e.g.,} $Conv~4\_3$ in SSD300) and its corresponding activation tensor $A\in{\mathbb{R}^{C\times{H}\times{W}}}$, which consists of $C$ feature planes and has height of $H$ and width of $W$. An intuitive idea for local feature alignment is to extract attention maps from both domains and somehow match them. However, the target attention map tends to focus on the predominant foreground objects instead of the full foreground objects (cf. Fig.~\ref{fig3}), which will impair the localization ability of detector for detecting those small or/and obscured objects. 
Thus, we resort to leverage classification output of the detection head (cf. Fig.~\ref{fig2}), which is denoted by $\hat{p}_m$ ($\hat{p}_m\in\mathbb{R}^{K+1}$, $m$ is the anchor index in $A$, and $m=\{1,2,...,H\times{W}\}$), to guide the object pattern matching. Specifically, the classification output $\hat{p}_m$ and the feature representation $A_m$ ($A_m\in\mathbb{R}^C$) are nonlinearly fused via tensor product operation, \emph{i.e.,} $\hat{A}_m={A_m\otimes\hat{p}_m}$, where $\hat{A}_m$ is the fused feature vector. In order to prevent the dimension explosion, we draw motivations from the randomized multilinear map~\cite{kar2012random,long2018conditional} to estimate the tensor product via Hadamard product, 
\begin{equation}\label{eq:estimation}
\hat{A}_m={(\boldsymbol{R}_1{A_m})\odot(\boldsymbol{R}_2\hat{p}_m)},\;A_m\in\mathbb{R}^{\hat{C}}
\end{equation}
where $\odot$ denotes the Hadamard product. $\boldsymbol{R}_1$ and $\boldsymbol{R}_2$ are random matrices and each of their element follows uniform distribution with univariance. $\hat{C}$ is the feature dimension after fusion ($\hat{C}$ is set to 1024 in our experiments). Based on the category-guided activation tensor $\hat{A}$, we output a spatial attention map via an activation-based mapping function: $\mathcal{F}: \mathbb{R}^{\hat{C}\times{H}\times{W}}\rightarrow{\mathbb{R}^{H\times{W}}}$, which can be written as follows:
\begin{equation}\label{eq:att}
(\mathcal{F}(\hat{A}))_m = \sum\limits_{c=1}^{\hat{C}}|\hat{A}_m^c|^2
\end{equation}
To reduce computational cost, we flatten the source and target attention maps to vectors, which are denoted as $f^s$ and $f^t$. Finally, We align the source and target object patterns by minimizing distance between the them, 
\begin{equation}\label{eq:copm}
	\mathcal{L}_{\text{la}} =\sqrt{H\times{W}}\cdot\Phi({\frac{f^s}{\left\Vert{f^s}\right\Vert}_2,\frac{f^t}{\left\Vert{f^t}\right\Vert}_2})
\end{equation}
where $\Phi(x, x')=\left\Vert{x-x'}\right\Vert_2$ is the Euclidean distance. Note that we incorporate a pixel-level domain discriminator~(\emph{i.e.,} $D_l$ in Fig.~\ref{fig2}) into COPM to further reduce the low-level feature disparity. Thus, the objective of COPM is formulated as: $\mathcal{L}_{\text{copm}}=\mathcal{L}_{\text{la}}+\mathcal{L}_{\text{adv}}$, where $\mathcal{L}_{\text{adv}}$ is a vanilla pixel-wise domain adversarial training loss.


\subsection{Regularized Joint Category Alignment}
Prototype\footnote{Prototype is the mean feature of the samples within the same class.}-based feature alignment has been widely explored to measure the category-level feature discrepancy in UDA~\cite{xie2018learning,Chen_2019_CVPR,Pan_2019_CVPR} and been applied to the two-stage cross-domain detection~\cite{zheng2020cross,xu2020cross}. 
However, considering the dense prediction property of one-stage detectors, prototype alignment may be error-prone in this case compared to adapting two-stage detectors where most negative proposals will be filtered out. 
Moreover, prior efforts only implement the prototype alignment in a certain high-level feature layer without considering the potential complementary effect of different domain-specific layers. 
Motivated by this, we propose a Regularized Joint Category Alignment (RJCA) module to achieve the category alignment at different domain-specific layers 
and regularize the average prediction consistency of different layers with respect to the same category.

In the light of fully convolutional and multi-level prediction characteristics of one-stage detectors, we aims at jointly enforcing the cross-domain category alignment in different layers. First of all, assume that the deep networks will generate the activations in different layers as $\{(z_i^{s1},...,z_i^{s|L|})\}_{i=1}^{N_s}$ and $\{(z_j^{t1},...,z_j^{t|L|})\}_{i=1}^{N_t}$, where $l\in{L}$ and $z\in{\mathbb{R}^{C\times{H}\times{W}}}$. Then, we resort to the per-pixel prediction to compute the prototype of each source class in layer $l$, which can be written as:
\begin{equation}\label{eq:prototype}
\bar{Z}_k^{s|l|}=\frac{1}{n_s^k}\sum\limits_{i=1}^{N_s}\sum\limits_{m=1}^{H\times{W}}y_s^{imk}\cdot{z_{im}^{s|l|}}
\end{equation}
where $n_s^k$ denotes the number of source objects labeled with class $k$, $m$ is the pixel index in $z$. $y_s^{imk}\in\{0,1\}$ is an indicator for determining whether the current pixel is predicted as class $k$. 
The source global prototype of each class is computed at the beginning of training.  
Let the prediction of detection head \emph{w.r.t.} a target object be represented by $\hat{p}(z_{jm}^{t|l|})$. 
The target local prototype is computed by:
\begin{equation}\label{eq:prototype}
\small
\bar{z}_k^{t|l|}=\frac{1}{\hat{n}_t^k}\sum\limits_{j=1}^{|B_t|}\sum\limits_{m=1}^{H\times{W}}y_t^{jmk}\cdot{z_{jm}^{t|l|}}
\end{equation}
where $\hat{n}_t^k$ denotes the number of objects that are assigned with pseudo label $k$ and $B_t$ is the mini-batch samples of the target domain. Similarly, we can obtain a set of source local prototypes $\{\bar{z}_k^{s|l|}\}_{k=1}^K$. The objective function of joint category alignment is formulated as follows:
\begin{equation}\label{eq:jca}
\footnotesize
\mathcal{L}_{\text{jca}}=\sum\limits_{l}[\underbrace{\sum\limits_{k}d({\bar{Z}_k^{s|l|}, \bar{Z}_k^{t|l|}})}_{\text{Compactness}}+\underbrace{\sum\limits_{m,n|m\neq{n}}h({\bar{Z}_m^{s|l|}, \bar{Z}_n^{t|l|}})}_{\text{Separation}}]
\end{equation}
where $d$ and $h$ are two different similarity functions to measure the distance between prototypes. In our case, we instantiate Eq.~\eqref{eq:jca} by the contrastive loss as defined in~\cite{hadsell2006dimensionality}. During training, the global prototype in Eq.~\eqref{eq:jca} is updated by the local prototype in a moving average manner,
\begin{equation}\label{eq:update}
\begin{split}
\small
\bar{Z}_k^{|l|}\leftarrow&\rho\bar{Z}_k^{|l|}+(1-\rho)\bar{z}_k^{|l|}
\end{split}
\end{equation}
where $\rho$ is set to $0.7$ in all experiments. In addition, we regularize the prediction consistency of different layers \emph{w.r.t.} the same class $k$ by respectively minimizing their symmetrized Kullback$-$Leibler (KL) divergence, which is formulated as:
\begin{equation}\label{eq:predction_distance}
\begin{split}
\mathcal{L}_{\text{pr}}=\frac{1}{K}\sum\limits_{l}\sum\limits_{k=1}^K\frac{1}{2}[D_{\text{KL}}(\hat{p}(\bar{z}_k^{t|l_a|})\Vert{\hat{p}(\bar{z}_k^{t|l_b|})}) \\
+D_{\text{KL}}(\hat{p}(\bar{z}_k^{t|l_b|})\Vert{\hat{p}(\bar{z}_k^{t|l_a|})})], \text{where}\;l_a, l_b\in{L}.
\end{split}
\end{equation}
where 
$\hat{p}(\bar{z}_k^{t|l_a|})$ and $\hat{p}(\bar{z}_k^{t|l_b|})$ stand for the average prediction \emph{w.r.t.} the class $k$ in different layers. Here, to smooth the prediction, we add a temperature variate $T$ ($T=2$ in all experiments) to the softmax function.
To this end, the objective of the proposed RJCA can be written as: $\mathcal{L}_{\text{rjca}}=\mathcal{L}_{\text{jca}}+\gamma\mathcal{L}_{\text{pr}}$, where $\gamma$ is set to $0.1$ in all experiments.
\subsection{Training Loss}
Suppose that the detection loss is denoted as $\mathcal{L}_{\text{det}}$, which includes the classification and regression losses. 
Joint all the presented parts, the overall objective function of I$^3$Net is formulated as follows,
\begin{equation}\label{eq:all}
\mathcal{L}_{\text{I$^3$Net}}=\mathcal{L}_{\text{det}}+\lambda_1\mathcal{L}_{\text{dcbr}}+\lambda_2(\mathcal{L}_{\text{copm}}+\mathcal{L}_{\text{rjca}})
\end{equation}
where $\lambda_1$ and $\lambda_2$ are hyper-parameters for balancing different loss components. 
\section{Experiments}
\subsection{Datasets}
We conduct experiments based on \textbf{Pascal VOC}~\cite{everingham2010pascal}, \textbf{Clipart1k}, \textbf{Watercolor2k}, and \textbf{Comic2k}~\cite{inoue2018cross} datasets. Following the previous one-stage method~\cite{kim2019self}, we utilize the Pascal VOC2007-trainval and VOC2012-trainval datasets as the source domain, and Clipart1k, Watercolor2k, and Comic2k as the target domain respectively. The Pascal VOC~\cite{everingham2010pascal}, which is a real-world image dataset, contains 16,551 images with 20 distinct object categories. Clipart1k~\cite{inoue2018cross}, which is a graphical image dataset with complex backgounds, consists of 1K images and has the same 20 categories as Pascal VOC. We utilize all images of Clipart1k as the target domain for both training and testing. Watercolor2k and Comic2k~\cite{inoue2018cross} contain 2K images respectively (\emph{i.e.,} 1K as the train set and the other 1K as the test set). They share 6 identical categories with the Clipart1k dataset, \emph{i.e.,} bicycle, bird, cat, car, dog, and person. Following the prior practice~\cite{kim2019self}, we leverage the train set for training and the test set for evaluation. 

\begin{center}
\begin{table*}[thb]
\caption{Results of adapting PASCAL VOC to Clipart1k (\%). mAP is reported on Clipart1k.}\label{table1}
\vspace{-0.1cm}
\centering
\footnotesize
\qquad 
\setlength\tabcolsep{1.5pt}
\begin{tabular*}{\hsize}{@{}@{\extracolsep{\fill}}cccccccccccccccccccccc@{}}
\toprule
Methods & aero & bcycle & bird & boat & bottle & bus & car & cat & chair & cow & table & dog & hrs & bike & prsn & plnt & sheep & sofa & train & tv & mAP \\
\hline
Source Only~\cite{liu2016ssd} & 27.3 & 60.4 & 17.5 & 16.0 & 14.5 & 43.7 & 32.0 & 10.2 & 38.6 & 15.3 & 24.5 & 16.0 & 18.4 & 49.5 & 30.7 & 30.0 & 2.3 & 23.0 & 35.1 & 29.9 & 26.7 \\
DANN~\cite{ganin2016domain} & 24.1 & 52.6 & 27.5 & 18.5 & 20.3 & 59.3 & 37.4 & 3.8 & 35.1 & 32.6 & 23.9 & 13.8 & 22.5 & 50.9 & 49.9 & 36.3 & 11.6 & 31.3 & 48.0 & 35.8 & 31.8 \\
DT+PL w/o label~\cite{inoue2018cross} & 16.8 & 53.7 & 19.7 & \textbf{31.9} & 21.3 & 39.3 & 39.8 & 2.2 & \textbf{42.7} & \textbf{46.3} & 24.5 & 13.0 & \textbf{42.8} & 50.4 & 53.3 & 38.5 & 14.9 & 25.1 & 41.5 & 37.3 & 32.7 \\
WST~\cite{kim2019self} & \textbf{30.8} & 65.5 & 18.7 & 23.0 & 24.9 & 57.5 & 40.2 & 10.9 & 38.0 & 25.9 & \textbf{36.0} & 15.6 & 22.6 & 66.8 & 52.1 & 35.3 & 1.0 & 34.6 & 38.1 & 39.4 & 33.8 \\
BSR~\cite{kim2019self} & 26.3 & 56.8 & 21.9 & 20.0 & 24.7 & 55.3 & 42.9 & 11.4 & 40.5 & 30.5 & 25.7 & 17.3 & 23.2 & 66.9 & 50.9 & 35.2 & 11.0 & 33.2 & 47.1 & 38.7 & 34.0 \\
SWDA$^\dagger$~\cite{saito2019strong} & 29.0 & 60.7 & 25.0 & 20.4 & 24.6 & 55.4 & 36.1 & 13.1 & 41.2 & 38.3 & 30.3 & 17.0 & 21.2 & 55.2 & 50.4 & 36.6 & 10.6 & 38.4 & 49.2 & 41.2 & 34.7 \\
BSR+WST~\cite{kim2019self} & 28.0 & 64.5 & 23.9 & 19.0 & 21.9 & \textbf{64.3} & \textbf{43.5} & 16.4 & 42.2 & 25.9 & 30.5 & 7.9 & 25.5 & 67.6 & 54.5 & 36.4 & 10.3 & 31.2 & \textbf{57.4} & 43.5 & 35.7 \\
HTCN$^\dagger$~\cite{chen2020harmonizing} & 28.7 & 67.7 & 25.3 & 16.1 & 28.7 & 56.0 & 38.9 & 12.5 & 41.0 & 33.0 & 29.6 & 12.9 & 22.9 & 69.0 & \textbf{55.9} & 36.1 & 11.8 & 34.1 & 48.8 & 46.8 & 35.8 \\
\hline
I$^3$Net w/o DCBR & 30.5 & 66.9 & 25.6 & 17.9 & 24.0 & 47.8 & 35.7 & 13.8 & 40.6 & 36.3 & 27.8 & 16.5 & 24.5 & \textbf{71.4} & 56.6 & 38.2 & 10.5 & \textbf{39.9} & 50.7 & 44.5 & 36.0 \\
I$^3$Net w/o COPM & 28.7 & 66.8 & 28.4 & 23.1 & 25.3 & 58.4 & 42.8 & \textbf{19.2} & 40.4 & 33.6 & 32.7 & 18.1 & 23.5 & 53.8 & 52.5 & 35.6 & 13.4 & 37.3 & 52.4 & 46.0 & 36.6 \\
I$^3$Net w/o RJCA & 28.8 & \textbf{67.8} & 25.4 & 16.2 & 28.9 & 56.1 & 39.0 & 12.6 & 41.1 & 33.1 & 29.7 & 13.0 & 22.9 & 69.1 & \textbf{55.9} & 36.3 & 11.9 & 34.2 & 48.9 & \textbf{46.9} & 35.9 \\
I$^3$Net (Full) & 30.0 & 67.0 & \textbf{32.5} & 21.8 & \textbf{29.2} & 62.5 & 41.3 & 11.6 & 37.1 & 39.4 & 27.4 & \textbf{19.3} & 25.0 & 67.4 & 55.2 & \textbf{42.9} & \textbf{19.5} & 36.2 & 50.7 & 39.3 & \textbf{37.8}  \\
\bottomrule
\vspace{-0.5cm}
\end{tabular*}
\end{table*}
\end{center}
\vspace{-0.9cm}
\subsection{Implementation Details}
The base detection model in our experiments follows the same setting in~\cite{inoue2018cross,kim2019self} that utilize SSD300~\cite{liu2016ssd} framework with VGG-16~\cite{simonyan2014very} architectures. The parameters of VGG-16 is fine-tuned from the model that has been pre-trained on ImageNet. In all experiments, the input images are resized to 300~$\times$~300 and we conduct all augmentations used in~\cite{liu2016ssd,kim2019self}. 
The batch size is selected as 32 (16 source images and 16 target images) to fit the GPU memory. We evaluate the cross-domain detection performance by reporting mean average precision (mAP) with a IoU threshold of 0.5 on the target domain. 
We adopt the stochastic gradient descent (SGD) optimizer for the detection network training with a momentum of 0.9, an initial learning rate of 0.001, weight decay of 5$\times{10^{-4}}$. The learning rate is decreased to 0.0001 after 50 epochs. 
Note that the multi-label classifier $F_{mlc}$ is pre-trained on the label source domain and keeps fixed when training our adaptation network.
Without specific notation, we set $\tau=0.5$ in Eq.~\eqref{eq:w1} and $\theta=0.5$ in Eq.~\eqref{eq:w}.
For the $L$ in RJCA, we set $L=\{Conv7, Conv9\_2\}$ for the I$^3$Net model based on SSD.
We set $\lambda_1=0.05$ and $\lambda_2=1$ in Eq.~\eqref{eq:all} for all experiments. Our experiments are implemented with the Pytorch deep learning framework.
\vspace{-0.05cm}
\subsection{Comparisons with State-of-the-Arts}
\paragraph{State-of-the-arts.} We make comparison to the state-of-the-art cross-domain object detection methods, including Domain Adversarial Neural Networks~\textbf{(DANN)}~\cite{ganin2016domain}, adversarial Background Score Regularization + Weak Self-Training~\textbf{(BSR+WST)}~\cite{kim2019self}, Strong-Weak Distribution Alignment~\textbf{(SWDA$^\dagger$)}~\cite{saito2019strong}, and Hierarchical Transferability Calibration Network~\textbf{(HTCN$^\dagger$)}~\cite{chen2020harmonizing}. The quantitative results of DANN, BSR, WST, and BSR+WST are cited from the original paper~\cite{kim2019self}. We reproduce the complete SWDA model on our one-stage scenarios. Moreover, we remove the context-aware instance-level alignment component from the HTCN model and re-implement the rest modules in our experiments. Note that mainstream cross-domain detection methods (\emph{e.g.,} \cite{chen2018domain,zhu2019adapting,cai2019exploring,He_2019_ICCV,xu2020cross,zheng2020cross,xu2020exploring}) are tailored for two-stage detector and cannot be simply extended to one-stage-based experiments since they highly count on the region proposal mechanisms.
	\begin{table}[!t]
		\caption{Results on adaptation from Pascal VOC to Watercolor2k (\%). mAP is reported on the Watercolor2k test set.}\label{table2}
		\vspace{-0.3cm}
		\centering
		\small
		\qquad 
		\setlength\tabcolsep{1.5pt}
		\scalebox{0.95}{
			\begin{tabular*}{\hsize}{@{}@{\extracolsep{\fill}}cccccccc@{}}
				\toprule
				Methods & bike & bird & car & cat & dog & person & mAP \\
				\hline
				Source Only~\cite{liu2016ssd} & 77.5 & 46.1 & 44.6 & 30.0 & 26.0 & 58.6 & 47.1 \\
				DANN~\cite{ganin2016domain} & 73.4 & 41.0 & 32.4 & 28.6 & 22.1 & 51.4 & 41.5 \\
				BSR~\cite{kim2019self} & \textbf{82.8} & 43.2 & \textbf{49.8} & 29.6 & 27.6 & 58.4 & 48.6 \\
				WST~\cite{kim2019self} & 77.8 & 48.0 & 45.2 & 30.4 & 29.5 & 64.2 & 49.2 \\
				SWDA$^\dagger$~\cite{saito2019strong} & 73.9 & 48.6 & 44.3 & 36.2 & 31.7 & 62.1 & 49.5 \\
				BSR+WST~\cite{kim2019self} & 75.6 & 45.8 & 49.3 & 34.1 & 30.3 & 64.1 & 49.9 \\
				HTCN$^\dagger$~\cite{chen2020harmonizing} & 78.6 & 47.5 & 45.6 & 35.4 & 31.0 & 62.2 & 50.1 \\
				\hline
				I$^3$Net w/o DCBR & 78.7 & 49.2 & 42.6 & 37.4 & 32.4 & 62.5 & 50.5 \\
				I$^3$Net w/o COPM & 75.6 & 49.2 & 45.9 & \textbf{37.9} & 33.2 & 63.6 & 50.9 \\
				I$^3$Net w/o RJCA & 81.8 & 46.3 & 40.4 & 33.3 & \textbf{34.0} & 65.1 & 50.2 \\
				I$^3$Net (Full) & 81.1 & \textbf{49.3} & 46.2 & 35.0 & 31.9 & \textbf{65.7} & \textbf{51.5} \\
				\bottomrule
				\vspace{-0.5cm}
		\end{tabular*}}
	\end{table}
\paragraph{Results on Clipart1k.} Table~\ref{table1} displays the adaptation results on Pascal VOC~$\rightarrow$~Clipart1k. Source Only denotes that the baseline SSD is trained on the source domain and directly tested on the target domain without any adaptation. The proposed I$^3$Net significantly outperforms all the compared methods in terms of mAP and improves over state-of-the-art by +2.0\% (35.8\% to 37.8\%). It is noteworthy that all components of the proposed I$^3$Net are designed appropriately and when we remove any one of these components, the final performance will drop accordingly. 
\vspace{-0.4cm}
\paragraph{Results on Watercolor2k and Comic2k.} Results on the tasks of Pascal VOC~$\rightarrow$~Watercolor2k and Pascal VOC~$\rightarrow$~Comic2k are reported on Table~\ref{table2} and Table~\ref{table3} respectively. I$^3$Net achieves better performance on most object categories, indicating that I$^3$Net is capable of learning more transferable representations and scalable for different cross-domain detection scenarios. It is noteworthy that I$^3$Net substantially exhibits better adaptation performance on the challenging transfer task (27.8\% to 30.1\%), \emph{i.e.,} Pascal VOC~$\rightarrow$~Comic2k, where the domain discrepancy is substantially large between source and target data.  
\begin{center}
	\begin{table}[htb]
		\caption{Results on adaptation from Pascal VOC to Comic2k (\%). mAP is reported on the Comic2k test set.}\label{table3}
		\vspace{-0.3cm}
		\centering
		\small
		\qquad 
		\setlength\tabcolsep{1.5pt}
		\scalebox{0.95}{
			\begin{tabular*}{\hsize}{@{}@{\extracolsep{\fill}}cccccccc@{}}
				\toprule
				Methods & bike & bird & car & cat & dog & person & mAP \\
				\hline
				Source Only~\cite{liu2016ssd} & 43.3 & 9.4 & 23.6 & 9.8 & 10.9 & 34.2 & 21.9 \\
				DANN~\cite{ganin2016domain} & 33.3 & 11.3 & 19.7 & \textbf{13.4} & 19.6 & 37.4 & 22.5 \\
				BSR~\cite{kim2019self} & 45.2 & 15.8 & 26.3 & 9.9 & 15.8 & 39.7 & 25.5 \\
				WST~\cite{kim2019self} & 45.7 & 9.3 & 30.4 & 9.1 & 10.9 & 46.9 & 25.4 \\
				BSR+WST~\cite{kim2019self} & \textbf{50.6} & 13.6 & 31.0 & 7.5 & 16.4 & 41.4 & 26.8 \\
				SWDA$^\dagger$~\cite{saito2019strong} & 47.4 & 12.9 & 29.5 & 12.7 & 19.1 & 44.1 & 27.6 \\
				HTCN$^\dagger$~\cite{chen2020harmonizing} & 50.3 & 15.0 & 27.1 & 9.4 & 18.9 & 46.2 & 27.8 \\
				\hline
				I$^3$Net w/o DCBR & 44.2 & 14.0 & \textbf{35.1} & 6.5 & 19.3 & 51.7 & 28.5 \\
				I$^3$Net w/o COPM & 47.1 & 14.5 & 32.3 & 7.1 & \textbf{20.3} & \textbf{51.8} & 28.9 \\
				I$^3$Net w/o RJCA & 45.0 & 12.1 & 33.9 & 8.0 & 20.1 & 50.5 & 28.3 \\
				I$^3$Net (Full) & 47.5 & \textbf{19.9} & 33.2 & 11.4 & 19.4 & 49.1 & \textbf{30.1} \\
				\bottomrule
				\vspace{-0.3cm}
		\end{tabular*}}
	\end{table}
\end{center}

\begin{center}
	\begin{table}[htb]
		\caption{Ablation of I$^3$Net on three transfer tasks (\%).}\label{table4}
		\vspace{-0.3cm}
		\centering
		\small
		\qquad 
		\scalebox{0.95}{
			\begin{tabular*}{\hsize}{@{}@{\extracolsep{\fill}}cccc@{}}
				\toprule
				Source & \multicolumn{3}{c}{Pascal VOC} \\
				Target & Clipart1k & Watercolor2k & Comic2k \\ 
				\hline
				DCBR w/o Dynamic & 37.3 & 51.4 & 29.2 \\
				DCBR w/o CB & 37.1 & 51.0 & 29.3 \\
				COPM w/o C & 36.8 & 51.1 & 29.0 \\
				COPM w/ MMD & 34.9 & 48.4 & 27.0 \\
				COPM w/ Adv & 37.0 & 50.7 & 29.8 \\
				RJCA w/o J & 36.6 & 50.8 & 29.1 \\
				RJCA w/o PR & 37.4 & 51.5 & 29.4 \\
				\hline
				I$^3$Net (Full) & \textbf{37.8} & \textbf{51.8} & \textbf{30.1} \\
				\bottomrule
				\vspace{-0.6cm}
		\end{tabular*}}
	\end{table}
\end{center}

\begin{figure*}[!t]
\centering
\includegraphics[width=0.90\textwidth]{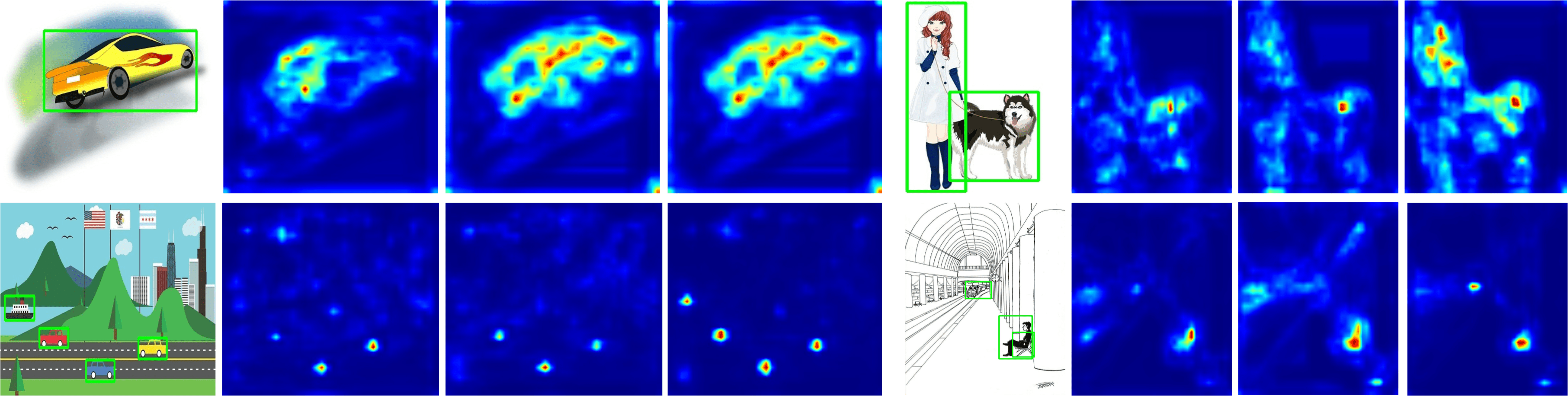}
\vspace{-0.15cm}
\caption{Illustration of the target attention maps generated by Source Only, HTCN$^\dagger$, and I$^3$Net. From left to right: input target images with ground-truth bounding boxes, Source Only, HTCN$^\dagger$, I$^3$Net.
}\label{fig3}
\end{figure*}

\begin{figure*}[!t]
\centering
\includegraphics[width=0.90\textwidth]{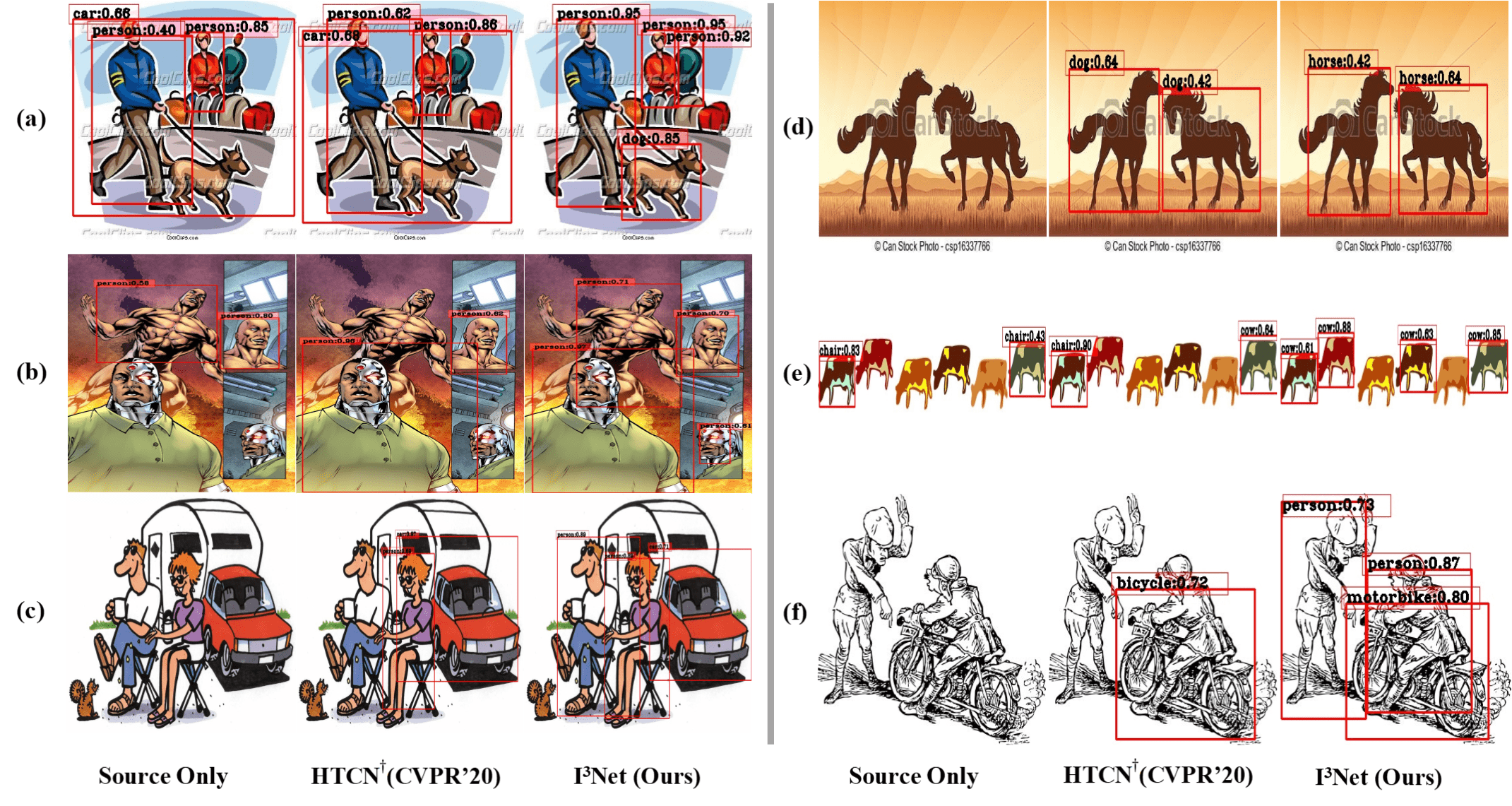}
\vspace{-0.15cm}
\caption{Qualitative detection results on Clipart1k, Watercolor2k, and Comic2k.}\label{fig4}
\vspace{-0.2cm}
\end{figure*}
\vspace{-1.5cm}
\subsection{Further Empirical Analysis}
\paragraph{Ablation Study.} We verify the effect of the proposed DCBR, COPM, and RJCA by evaluating variants of I$^3$Net. The results are reported in Table~\ref{table4}. \textbf{(1)} DCBR w/o Dynamic and DCBR w/o CB denote that we remove $w_1^t$ and $w_2^t$ from Eq.~\eqref{eq:w} respectively. \textbf{(2)} COPM w/o C denotes that we remove the non-linear fusion step (Eq.~\eqref{eq:estimation}) and directly match the source and target vectorized attention maps. COPM w/ MMD and COPM w/ Adv denote that we replace the $L_2$ distance in Eq.~\eqref{eq:copm} by MMD~\cite{long2015learning} and domain adversarial loss~\cite{ganin2015unsupervised} respectively. \textbf{(3)} 
RJCA w/o J is the variant that only conducts the category alignment in one layer. RJCA w/o PR is the variant without prediction regularization (Eq.~\eqref{eq:predction_distance}). The results of COPM w/ MMD and COPM w/ Adv reveal that $L_2$ distance is able to better preserve the structured information (\emph{i.e.,} object patterns). The results of RJCA w/o J verify the significance of considering the complementary effect of different domain-specific layers.       
\vspace{-0.5cm}
\paragraph{Visualization of COPM.} Figure~\ref{fig3} visualizes the attention maps generated by Source Only~\cite{liu2016ssd}, HTCN$^\dagger$~\cite{chen2020harmonizing}, and I$^3$Net (Ours). The brighter the color is, the larger the weight value is. It is notable that the proposed I$^3$Net is capable of (i) capturing the discriminative regions which contain rich semantic information, (ii) highlighting the foreground objects even with small object size, and (iii) suppressing the redundant background information.  
\vspace{-0.5cm}
\paragraph{Qualitative detection results.} Figure~\ref{fig4} demonstrates the example of detection results on the three target domains, \emph{i.e.,} Clipart1k, Watercolor2k, and Comic2k. The proposed I$^3$Net consistently and significantly outperforms both Source Only~\cite{liu2016ssd} and HTCN$^\dagger$~\cite{chen2020harmonizing} models in different transfer tasks. Owing to the introduction of DCBR, I$^3$Net is capable of precisely detecting the sample-scarce categories (\emph{e.g.,} (a), (d), and (e)). I$^3$Net is able to detect those obscured objects and provide accurate bounding box predictions since we explicitly encourage the alignment of cross-domain object patterns via the proposed COPM (\emph{e.g.,} (a), (b), (c), and (f)). In addition, due to the presence of RJCA, I$^3$Net is able to ensure the cross-domain semantic consistency, and thus significantly reduce the false positive results and enhance the classification accuracy (\emph{e.g.,} (d) and (e)). 
\vspace{-0.2cm}
\section{Conclusion}
In this paper, we proposed the Implicit Instance-Invariant Network (I$^3$Net) to solve the cross-domain object detection problem based on the one-stage detectors without requiring explicit instance-level features. 
The key idea of our method is to implicitly learn instance-invariant features via exploiting the natural characteristics of deep features in different layers,
\emph{i.e.,} suppressing redundant information from the lower layers and enhancing the cross-domain semantic correlation of foreground objects at the higher layers. 
Experiments on three standard cross-domain detection benchmarks verified the effectiveness of our method.
\vspace{-0.5cm}
\paragraph{Acknowledgement}
This work was partially supported by National Key Research and Development Program of China (No.2020YFC2003900) and the National Natural Science Foundation of China under Grants U19B2031, 61971369.

{\small
\bibliographystyle{ieee_fullname}
\bibliography{I3Net}
}
\end{document}